%% file: main.tex
\documentclass[runningheads]{llncs}

\usepackage{graphicx}
\usepackage{mathtools}
\usepackage{amsmath}
\usepackage{amssymb}
\usepackage{wrapfig}
\usepackage{booktabs}
\usepackage{multirow}
\usepackage{subcaption} 
\usepackage[table]{xcolor}

\begin{document}
\title{Learning to Match 2D Keypoints Across Preoperative MR and Intraoperative Ultrasound}

\titlerunning{MR-US Keypoints Matching}

\author{Hassan Rasheed\inst{1,3,4} \and
Reuben Dorent\inst{1} \and
Maximilian Fehrentz\inst{1,3} \and \\
Tina Kapur\inst{1} \and
William M. Wells III\inst{1,2} \and
Alexandra Golby\inst{1} \and
Sarah Frisken\inst{1} \and \\
Julia A. Schnabel\inst{3,4} \and
Nazim Haouchine\inst{1}}

\authorrunning{Rasheed et al.}


\institute{
Harvard Medical School, Brigham and Women's Hospital, Boston, MA, USA \and
Massachusetts Institute of Technology, Cambridge, MA, USA \and
Technical University of Munich, Munich, Germany \and
Helmholtz Center Munich, Munich, Germany}

%
\maketitle              
\begin{abstract}
We propose in this paper a texture-invariant 2D keypoints descriptor specifically designed for matching preoperative Magnetic Resonance (MR) images with intraoperative Ultrasound (US) images.
We introduce a \textit{matching-by-synthesis} strategy, where intraoperative US images are synthesized from MR images accounting for multiple MR modalities and intraoperative US variability.
We build our training set by enforcing keypoints localization over all images then train a patient-specific descriptor network that learns texture-invariant discriminant features in a supervised contrastive manner, leading to robust keypoints descriptors. 
Our experiments on real cases with ground truth show the effectiveness of the proposed approach, outperforming the state-of-the-art methods and achieving $80.35\%$ matching precision on average. 
\end{abstract}

\input{intro}
\input{method}
\input{results}

\section{Conclusion}
We presented a novel multimodal image matching method between preoperative MR and intraoperative US images.
Our matching-by-synthesis strategy, coupled with a patient-specific contrastive learning approach led to a texture-invariant descriptor, capable of matching 2D keypoints under different acquisition variations. 
Future work will extend the descriptor to 3D affine transformations and will integrate physics-based modelling~\cite{talbot2015}\cite{paulus2017handling} to account for elastic deformations and tissue resections, essential to achieve post-resection MR-US registration.

\subsubsection*{Acknowledgement.}
This work was supported by the National Institutes of Health grants R01EB032387, R01EB034223, R03EB033910, and K25EB035166.

\bibliographystyle{splncs04}
\bibliography{mybib}

\end{document}

%% file: intro.tex
\section{Introduction}
\label{sec:intro}
Multimodal image matching is a fundamental problem that involves identifying and pairing similar features or patterns across images from different modalities, with significant appearance changes \cite{jiang2021review}.
It has a wide range of applications in medical imaging, including image retrieval and classification \cite{jiang2021review,kumar2013content}, slice-to-volume alignment~\cite{ferrante2017slice} and image registration~\cite{evan2021keymorph,Machado,JIE,heinrich2012mind,joutard2022driving}.
When used during image-guided surgery, it can provide surgeons with complementary imaging information from various modalities, facilitating the identification of key anatomical and surgical structures for improved surgical outcomes.
For instance, during neurosurgery, intraoperative Ultrasound (US) is often used in conjunction with preoperative Magnetic Resonance Imaging (MRI) to localize tumor boundaries that may have been shifted due to brain shift \cite{Gonzalez-Darder2019}. 
This allows surgeons to achieve maximally safe resection, which is positively correlated with a patient's chances of survival \cite{Sanai}\cite{Haouchine2022}.
However, although affordable and real-time in comparison to intraoperative MRI, US images can be difficult to interpret \cite{dixon2022intraoperative}, requiring image registration with preoperative MRI to disambiguate US images.

In this work, we focus on keypoint-based multi-modal methods, where correspondences between a relatively small set of keypoints extracted from both images are first identified before being matched. 
They typically rely on discriminative descriptors that can be matched under various imaging conditions allowing for robust matching. 
This area has been extensively studied \cite{jiang2021review} and some approaches have been successfully applied to medical images, in MRI with different weights, such as T1, T2, and proton density, or with angiographic retinal images \cite{christy2014retinal}.
However, these methods are limited to preoperative images where dissimilarities between modalities are relatively small. 
MR-US image matching is a non-trivial task due to the large dissimilarity between these two modalities~\cite{wu2018multimodal}. 
Moreover, these modalities provide different textures, and volumetric information, operate at different spatial resolutions, and are corrupted by various sources of noise.
In particular, MRI uses pulse sequences to obtain images with contrasts between soft tissue types, producing high-resolution 3D volumetric images, whereas US acquires partial and noisy images that echo back structures based on wave distances.
Adapting such methods to MR-US images requires modeling the intraoperative texture gap related to US acquisitions.
A texture-invariant feature descriptor can take several forms \cite{jiang2021review} focusing on temporal changes \cite{verdie2015tilde}, structural changes \cite{baruch2021joint} or appearance changes \cite{detone2018superpoint}.

This work presents novel texture-invariant 2D keypoint descriptors designed explicitly for matching preoperative MR with intraoperative US images.
We introduce a \textit{matching-by-synthesis} strategy, where intraoperative US images are synthesized from a patient-specific MR image and then used to train a cross-modality descriptor network.
This network is trained in a supervised contrastive manner to be agnostic to US texture changes and to be robust to speckle noise. 
Our approach does not require human-annotated key points or a large training dataset.
Moreover, our method is interpretable since the matched and mismatched keypoints can be visualized.
Our experiments on real cases with ground truth show the effectiveness of the proposed approach, outperforming the state-of-the-art methods. 

%% file: method.tex
\begin{figure}[ht!]
     \begin{center}
        \includegraphics[width=0.99\linewidth]{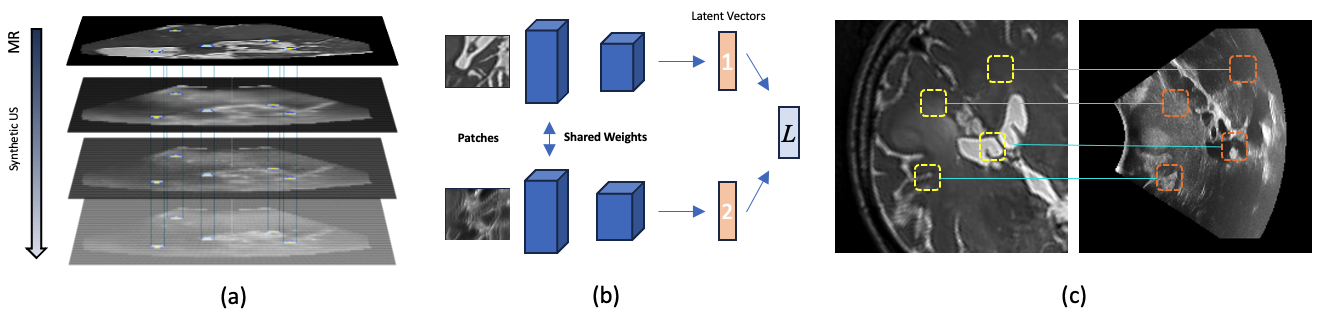}
    \end{center}
    \caption{Method overview: We rely on training images composed of one MR image and multiple synthesized US images, generated under different modes and noise levels (a). We train a Siamese network on image patches to learn similar and dissimilar features in a supervised contrastive manner (b). Applying this  network to patches from each image leads a MR-US cross-modal matching (c).}
    \label{fig:overview}
\end{figure} 

\section{Methods}
\subsection{Overview and Problem Setting}
Let us assume a preoperative MR image $\mathbf{I}_\text{MR}\in\mathbb{R}^{H\times W\times D}$, an intraoperative US image $\mathbf{I}_\text{US}\in\mathbb{R}^{H\times W\times D}$, and a set of 2D points of interest $\mathbf{x}  = \{x_i | i \in 1 \dots n \} \in \mathbf{I}_\text{MR}$ and $\mathbf{y}  = \{y_i | i \in 1 \dots m \} \in \mathbf{I}_\text{US}$, independently detected on each image.
We seek at finding a mapping $\pi : \{1,2, \dots, n\} \rightarrow \{1,2, \dots, m\}$ which maximizes the similarity function $\mathcal{M}$:
\begin{equation}
\operatorname*{argmax}_{\pi \in \Pi(n)} \mathcal{M} \big( [\mathbf{d}(x_{\pi(i)} )]^n_{i=1}, [\mathbf{d}(y_i)]^m_{i=1} \big)
\end{equation}
where $\Pi(n)$ denotes all possible mappings and $\mathbf{d}$ is a 2D keypoint descriptor.

To build the cross-modality descriptor $\mathbf{d}$, we train a descriptor network on image patches by minimizing the sum of the loss for pairs of corresponding and non-corresponding patches in a supervised contrastive manner. 
Since this type of training requires a large number of paired images $(\mathbf{I}_\text{MR}, \mathbf{I}_\text{US})$ to efficiently mine positive and negative patches, we propose a \textit{matching-by-synthesis} strategy, where intraoperative US images are synthesized from preoperative MR images using a generative network.
We describe below how we build the cross-modality training dataset and train the 2D descriptor network (See Fig~\ref{fig:overview}).

\subsection{Intraoperative Image Synthesis}
\begin{figure}[t]
     \begin{center}
        \includegraphics[width=1\linewidth]{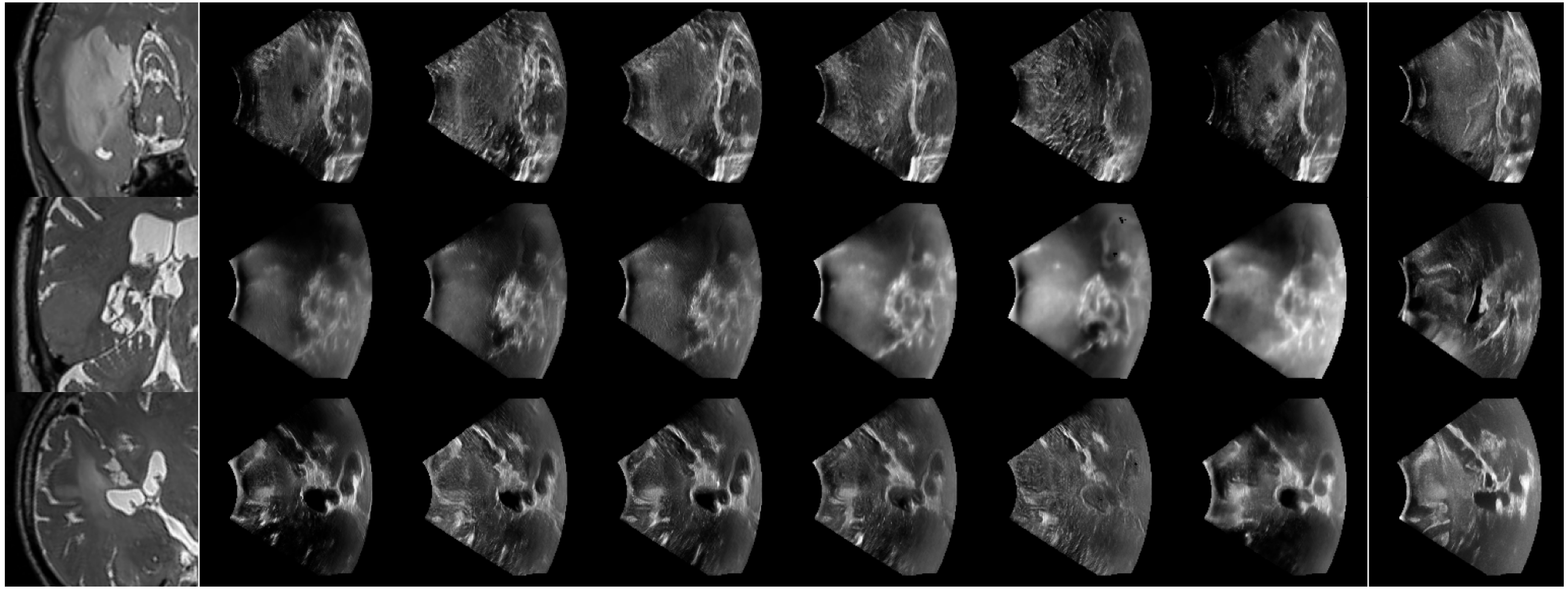}
    \end{center}
    \caption{Synthetic US image generations for three different T2 MR images (One case per row). The first column shows T2 MR; the middle columns show samples of synthetic US images generated using different combinations of T2, T1, and FLAIR with different speckles; the last column shows the ground truth US image.}
    \label{fig:synth}
\end{figure}

To synthesize an intraoperative US image $\mathbf{I}_\text{SynUS}$ from a preoperative MR image $\mathbf{I}_\text{MR}$, we define a generative network $g(\cdot)$ so that $\mathbf{I}_\text{SynUS} = g(\mathbf{I}_\text{MR}, \widehat{\theta}_g, \gamma)$, with $\widehat{\theta}_g$ being the network pre-trained parameters and $\gamma$ are a set of parameters to vary the texture of the generated image at inference. 
We rely on the multimodal hierarchical variational auto-encoder (MHVAE) proposed in \cite{dorent2023unified}, which is the current state-of-the-art for MR to iUS synthesis. MHVAE has the flexibility to handle incomplete sets of MR images as input and produces realistic US synthesis (See Fig~\ref{fig:synth}), allowing us  to synthesise ultrasound for any combination of input modalities (T1, T2, FLAIR MRs). Moreover, this method uses a principled probabilistic fusion operation to create a common representation space between modalities and a hierarchical latent structure to represent global features with the coarsest latent variable while the finer variables capture local characteristics. Sampling is performed at each level of the hierarchy to perform synthesis. By varying $\gamma$, the set of sampling parameters, we can generate images US images with different speckles and content from any combination of input modalities (T1, T2, FLAIR MRs). This allows us to create a 1-to-many set of paired images $T = \{\mathbf{I}_\text{MR}, \mathbf{I}_{\text{SynUS}}^1, \dots,\mathbf{I}_{\text{SynUS}}^p\}$ that will be used to build the training dataset.
We generate 28 synthetic US images ($p=28$) for each MR with the following combination: T1, T2, FLAIR, T1+T2, T1+FLAIR, T2+FLAIR, and T1+T2+FLAIR with 4 different sampling parameters.

\subsection{Building the Training Dataset}
As in \cite{baruch2021joint}, we chose to train our network on 2D patches, that we denote $\mathbf{p}$, extracted at keypoints locations, and standardized into $s\times s$ pixels. 
We use SuperPoint \cite{detone2018superpoint}. Note that other detectors could be considered as well. 
To collect a training set of positive patches, we first detect keypoints on the MR image $\mathbf{I}_\text{MR}$.
We then iterate over the synthesized US images from $\{\mathbf{I}_{\text{SynUS}}^{i}\}_{i=1}^p$ and enforce SuperPoint to detect keypoints at the same location as keypoints from $\mathbf{I}_\text{MR}$. 
If a keypoint is detected at the same location (within a 5px margin) in at least 3 images, its location is likely to be a good candidate to learn.
We then cluster keypoints within a 5px radius using the DBSCAN method.
We normalize the patches with the grayscale mean and standard deviation of the entire training set.
We choose a patch size of $s=64$ pixel and extract $256$ patches per slice while retaining about half after clustering. 

This strategy allows us to keep only the most repeatable keypoints for training, discarding the ones that were detected only on one modality. 
Moreover, our network is trained to learn texture-invariance by enforcing keypoints locations over all sets of synthetic US images.

\subsection{Learning Cross-modal Feature Descriptor}
\subsubsection{Model Architecture and Loss Function}
Given an image patch $\mathbf{p}$ around a detected keypoint, the objective is to build a descriptor $\mathbf{d}$ that retains structural similarity between modalities but is invariant to changes in textures. 
Following on related work, our descriptor network is a convolutional neural network $h(\cdot)$ with a relatively simple Siamese architecture so that $\mathbf{d} = h(\mathbf{p}, \theta_h)$, where $\theta_h$ are the network parameters.
To train $h(\cdot)$, we consider corresponding and non-corresponding
pairs of patches and use a \textit{Triplet Loss} that learns the ideal embedding space for the patches. 
We propose to pair patch $\mathbf{p}^k_a \in \mathbf{I}_\text{MR}$ with multiple positive and negative patches $\mathbf{p}^k_p \in \mathbf{I}_\text{SynUS}$ and $\mathbf{p}^{k'}_{n} \in \mathbf{I}_\text{SynUS}$ respectively (with $k \neq k'$), enforcing the network to discard texture changes between modalities while learning content similarity. 
Formally, the loss is defined as:
\begin{equation}
    L_{triplet}(\mathbf{d}^k_a, \mathbf{d}^k_p, \mathbf{d}^{k'}_n) = \sum_k \text{max}\big(0, | \mathbf{d}^k_a - \mathbf{d}^k_p|^2 - | \mathbf{d}^l_a - \mathbf{d}^{k'}_n|^2 + C\big)
\end{equation}
where $C=1$ is the margin.
We use hard mining during training, which was shown in \cite{baruch2021joint} to be critical for descriptor performance. 
we select the negative samples from within the training batch that have the lowest loss, based on $L_2$ distance, against the current patch and use that for backpropagation.
We use balanced batches of positive and negative pairs.

\subsubsection{Training and Optimization Details}
We train our model in 2D, pairing individual slices rather than the whole volume. 
This makes the learning scalable without loss of information, as most volume regions do not contain keypoints. 
We include all the slices in the training.  
We train one model per patient, in a patient-specific manner.
This also allows us to maintain a batch with multiple patch pairs, which helps
convergence. 
The training typically takes less than 30mn on a single 10GB GPU.
For optimization, we use ADAM with a learning rate of $10^{-3}$ and a batch size of $256$.

\subsubsection{Run-time Inference and Matching}
At inference, we set the detection to $n=200$ keypoints from the MR image. For the US image, rather than specifying an exact number of keypoints, $m$, we impose a limit of $1500$ keypoints. 
We build the descriptors on each patch by running a feed-forward inference using one branch of the Siamese network.
Our similarity function $\mathcal{M}$ takes the form of a K nearest neighbors (KNN) similarity search via Cosine-based representations, which we found empirically to perform better than an $L_2$ similarity. 
We use the standard criteria consisting of distance threshold, Lowe's ratio test, and matching uniqueness to filter out false negative matches.

\begin{figure}[h!]
\centering 
 \subfloat{\includegraphics[width=0.33\linewidth]{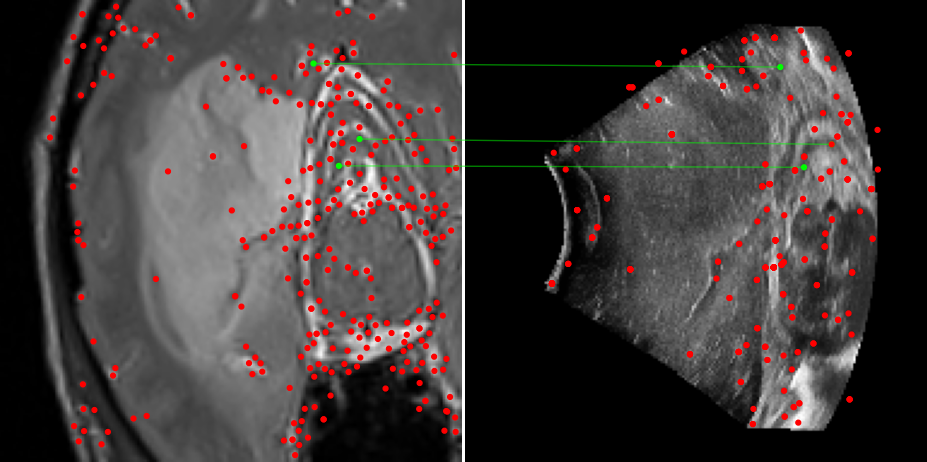}}
 \hfill
 \subfloat{\includegraphics[width=0.33\linewidth]{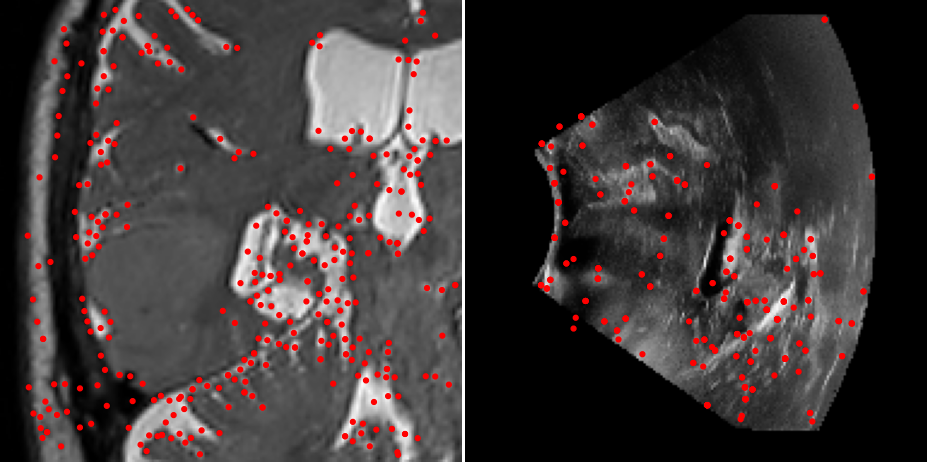}}
 \hfill
 \subfloat{\includegraphics[width=0.33\linewidth]{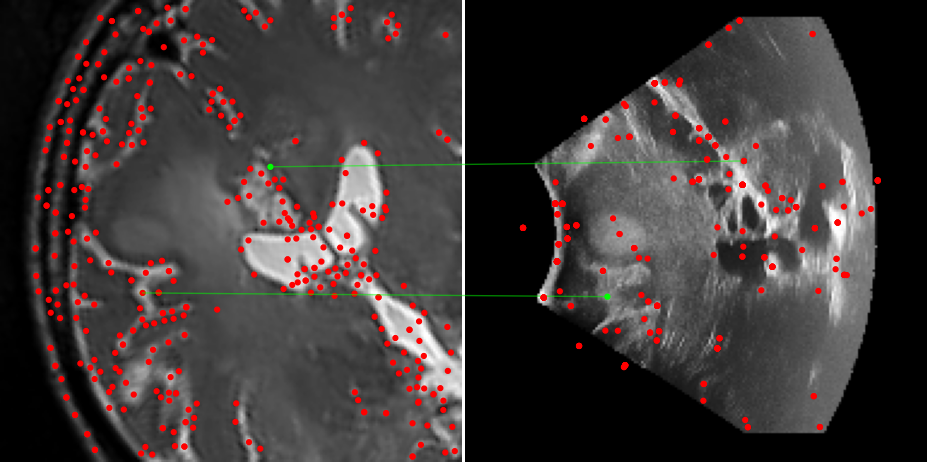}} \\

 \subfloat{\includegraphics[width=0.33\linewidth]{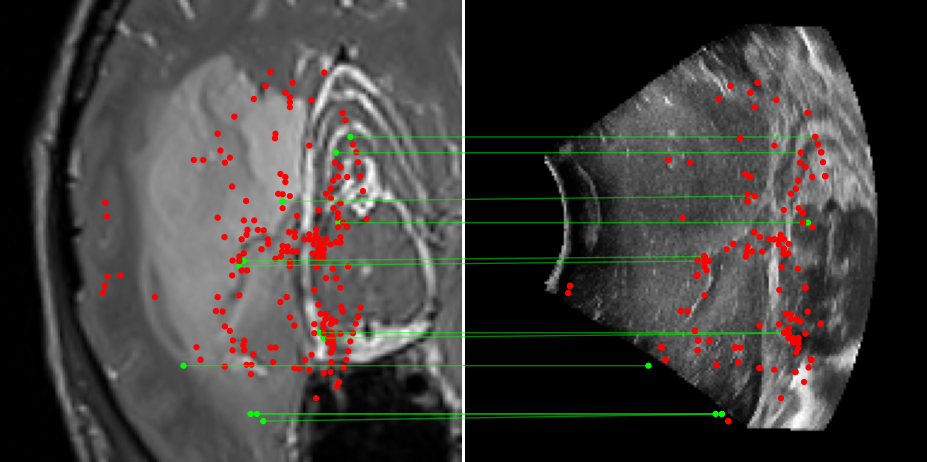}}
 \hfill
 \subfloat{\includegraphics[width=0.33\linewidth]{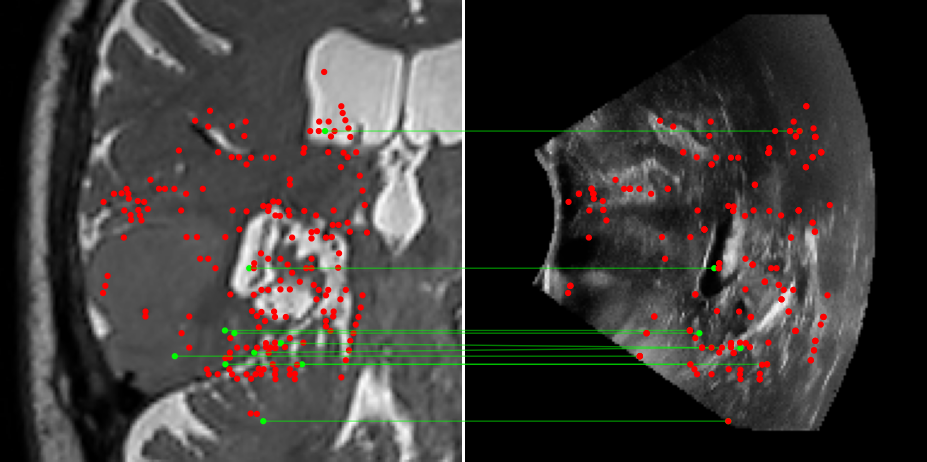}}
 \hfill
 \subfloat{\includegraphics[width=0.33\linewidth]{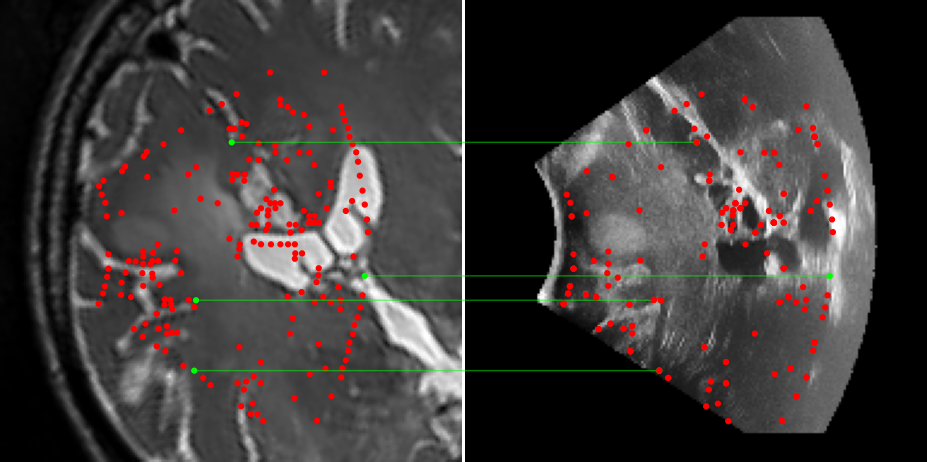}} \\

 \subfloat{\includegraphics[width=0.33\linewidth]{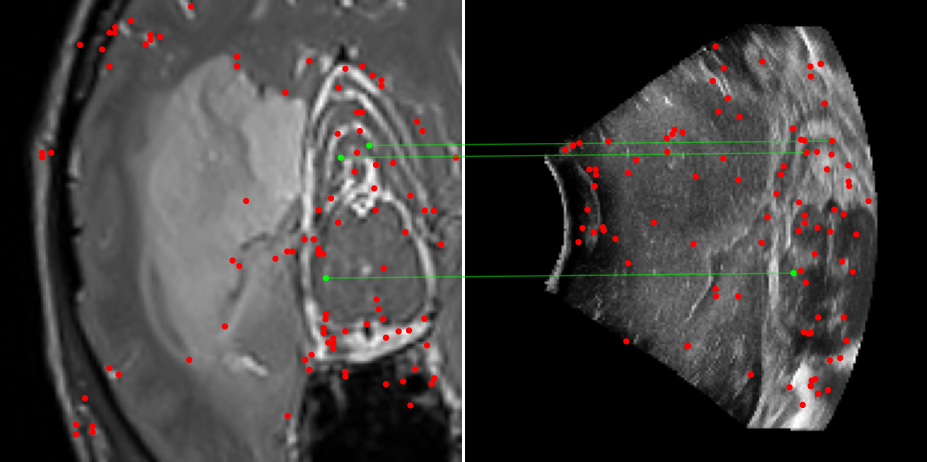}}
 \hfill
 \subfloat{\includegraphics[width=0.33\linewidth]{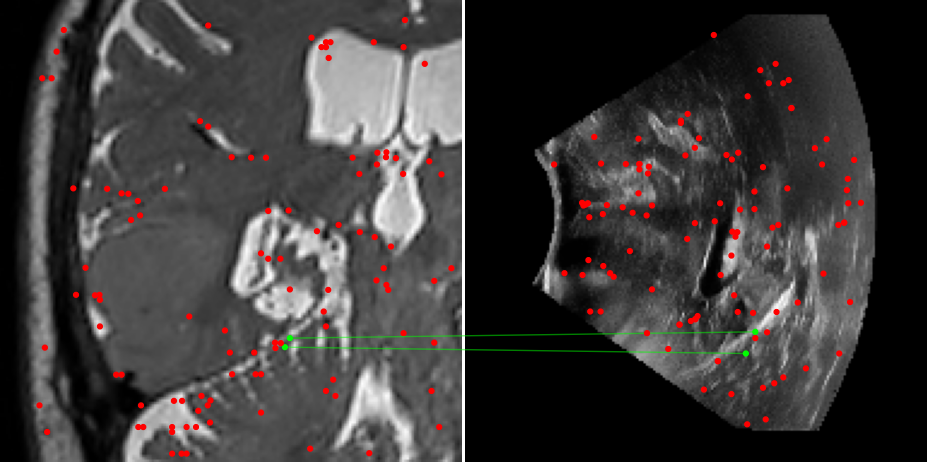}}
 \hfill
 \subfloat{\includegraphics[width=0.33\linewidth]{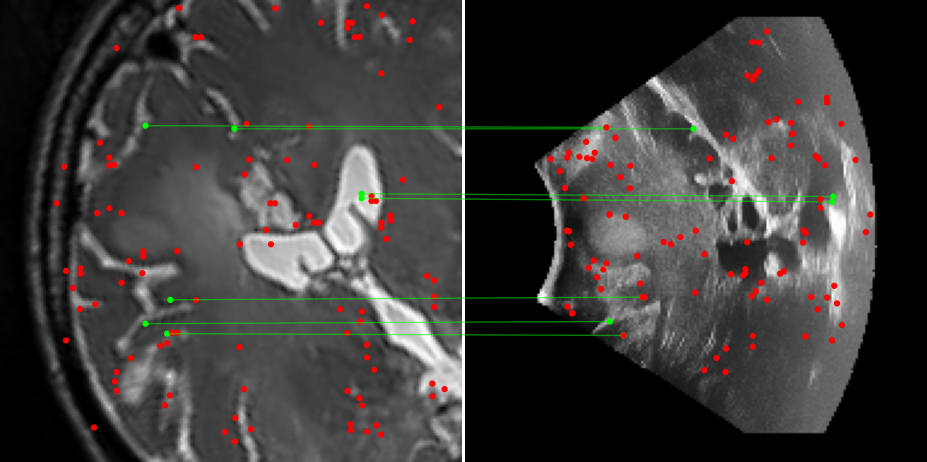}} \\
 
 \subfloat{\includegraphics[width=0.33\linewidth]{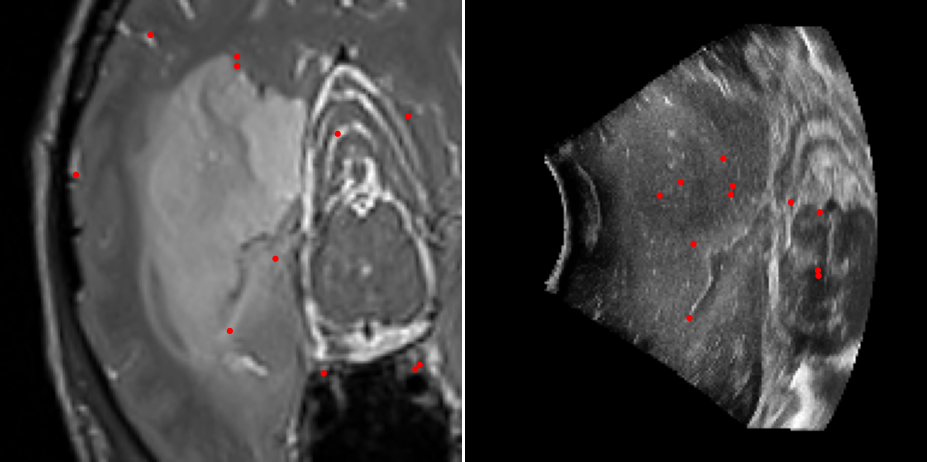}}
 \hfill
 \subfloat{\includegraphics[width=0.33\linewidth]{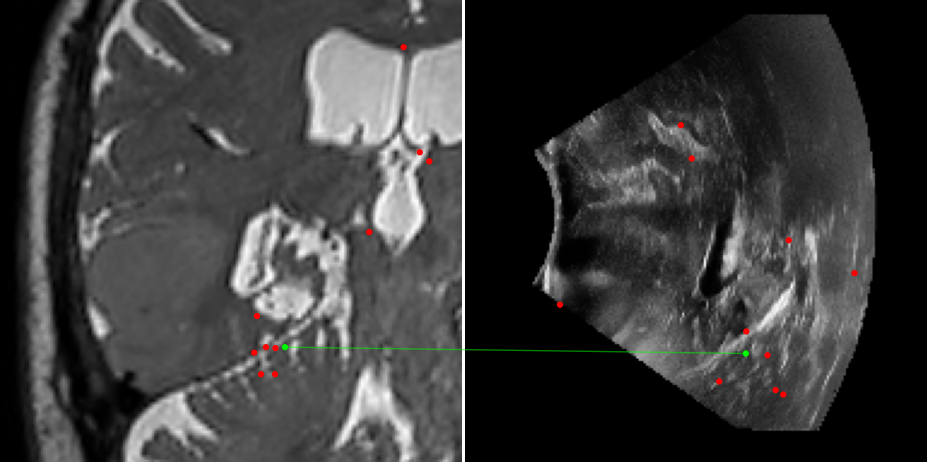}}
 \hfill
 \subfloat{\includegraphics[width=0.33\linewidth]{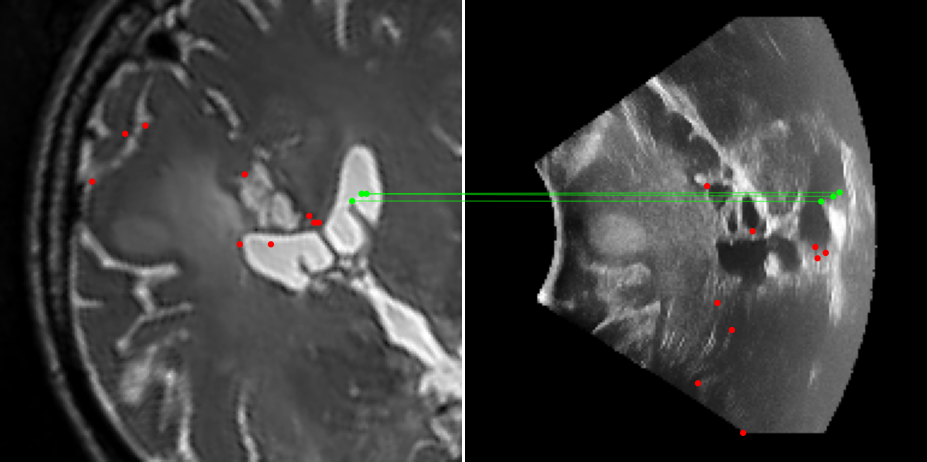}} \\

 \subfloat{\includegraphics[width=0.33\linewidth]{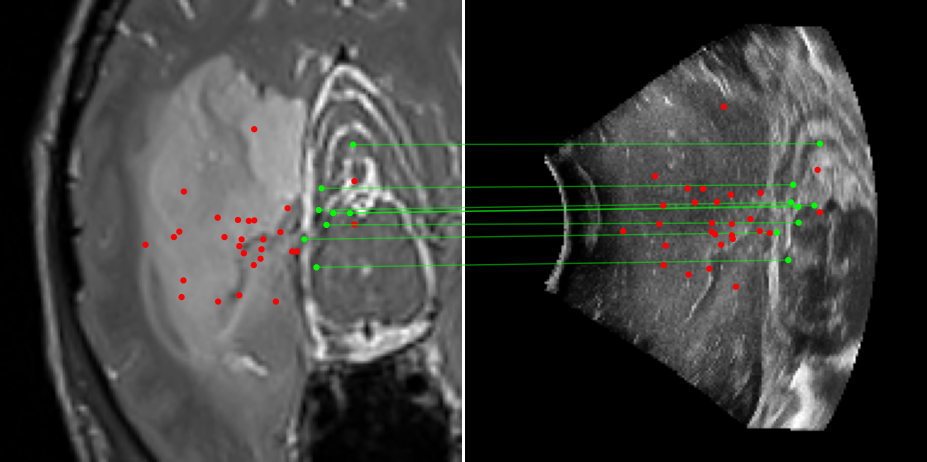}}
 \hfill
 \subfloat{\includegraphics[width=0.33\linewidth]{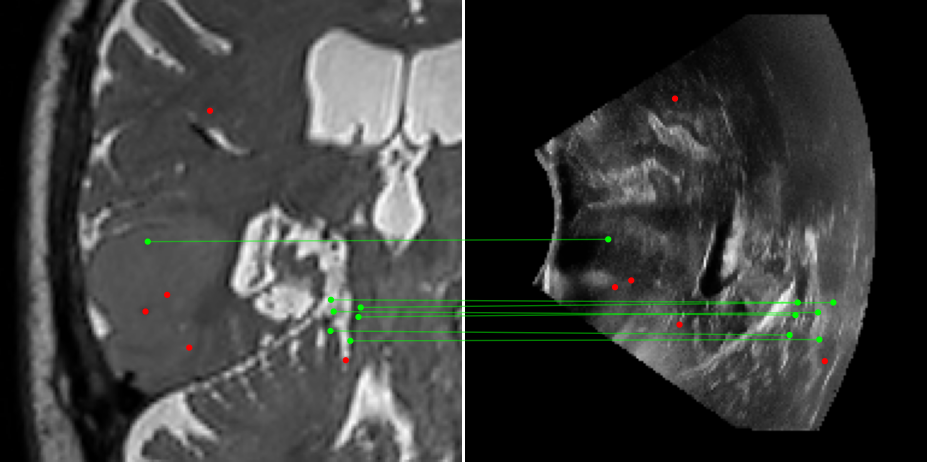}}
 \hfill
 \subfloat{\includegraphics[width=0.33\linewidth]{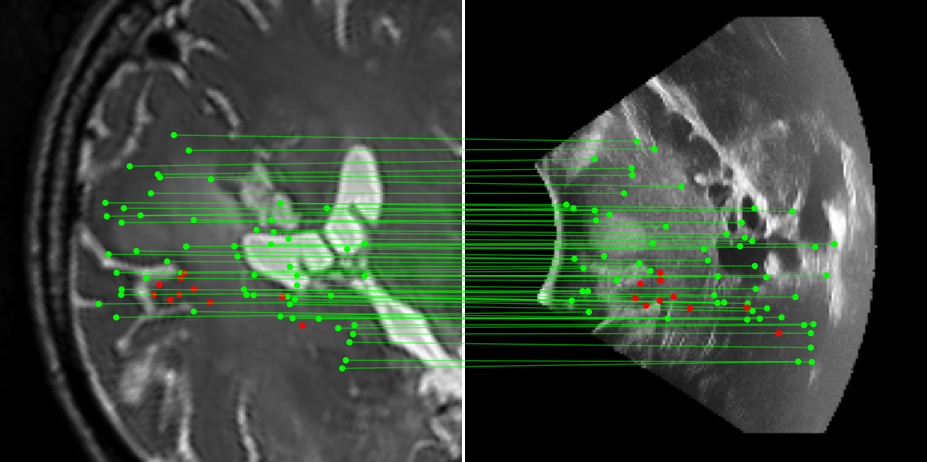}} \\

 \subfloat{\includegraphics[width=0.33\linewidth]{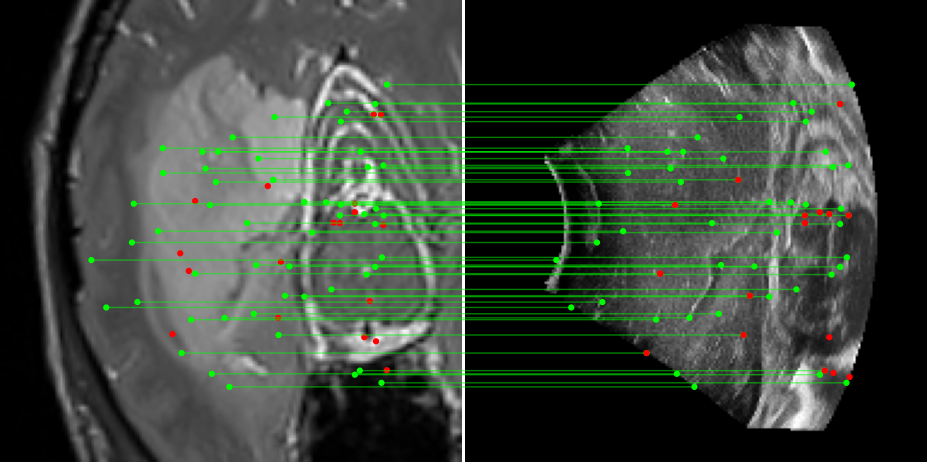}}
 \hfill
 \subfloat{\includegraphics[width=0.33\linewidth]{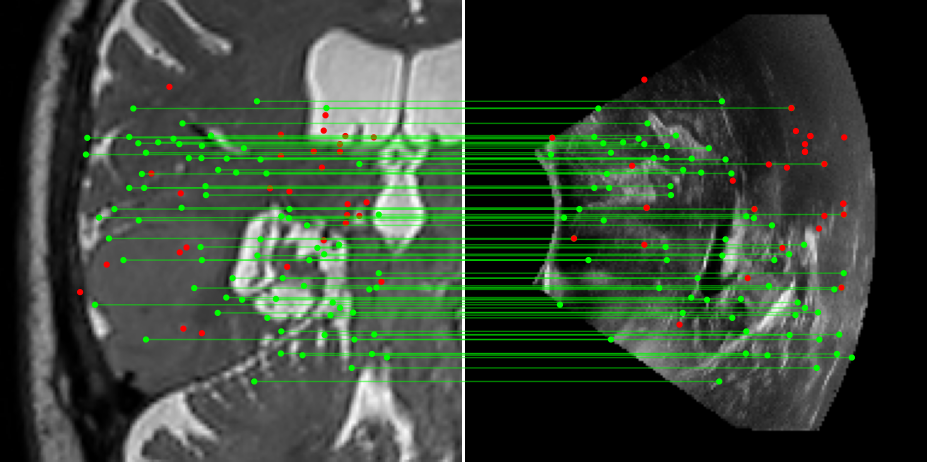}}
 \hfill
 \subfloat{\includegraphics[width=0.33\linewidth]{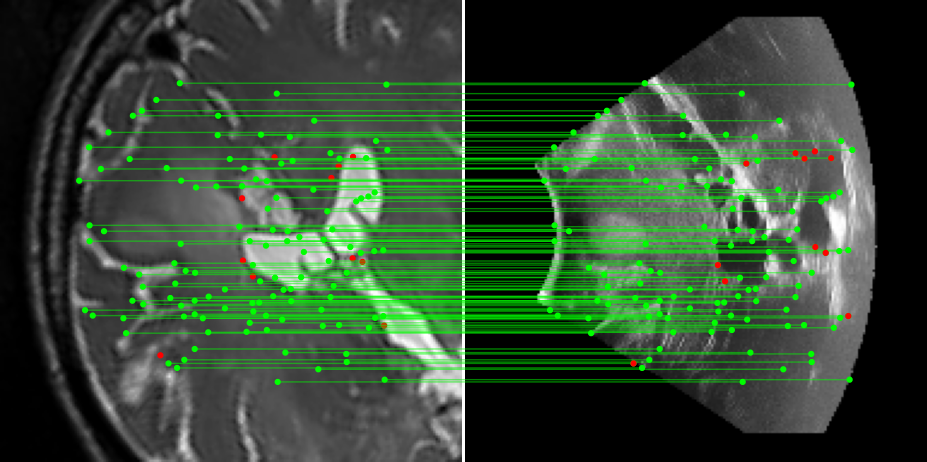}} \\

\caption{Examples of matching on three cases, one per column (MR on left and US on right). From top to bottom: SIFT+Cosine, MIND+Cosine, SP+Cosine, SP+LG, Ours+LG, Ours+Cosine. Correct matches recovered by each method are shown in green lines and mismatched are shown with a red dot.}
\label{fig:matching}
\end{figure}

%% file: results.tex
\section{Experimentations and Results}
\subsubsection{Data.}
We evaluate our method on a dataset of 7 cases where both pre-operative 3D T2-SPACE and pre-dural opening intraoperative US reconstructed from a tracked handheld 2D probe were acquired. 
We used the ReMIND dataset \cite{juvekar2023remind} where 3D T2-SPACE scans are registered with the pre-dural US. 
Images are resampled to an isotropic $0.5\text{mm}$ resolution, padded for an in-plane matrix of $(192,192)$, and normalized in $[-1,1]$.

\subsubsection{Metrics.}  
Since paired data with ground truth is available for evaluation, we use keypoints locations (within a 4px radius) to evaluate our method.
We report the following metrics: Matching Score (MSc.) as the ratio of ground truth correspondences over the number of detected keypoints of the whole pipeline and Precision (Prec.) as the ratio of ground truth correspondences over the number of matched keypoints. We also report the number of matched points {MP}. 
Overall, our method achieves an average matching score of 26.62\%, an average matching precision of 80.35\%, and an average of 43.33 matched points.

\def\arraystretch{0.8}
\rowcolors{2}{}{gray!10}
\begin{table*}[t!]
\centering
\caption{Impact of modalities synthesis (Averages over $\approx$ 80 slices)}
\begin{tabular}{p{2em} p{2em} p{2em} | c | c | c | c }
\toprule
 \multicolumn{3}{c|}{T2 T1 FLAIR} & Prec. (\%) \hfill & MSc. (\%) \hfill &  Avg MP \hfill & Area (\%)\\
 \midrule 
 \, $\bullet$ & $\circ$ & $\circ$ &  \textbf{85.64}    &   7.06    &  16.50    &   25.05    \\  
 \, $\bullet$ & $\circ$ & $\bullet$ &  83.01    &   7.60    &  18.33    &   30.73    \\  
 \, $\bullet$ & $\bullet$ & $\circ$ &  83.25    &   12.87    &  30.92    &   44.77    \\  
 \, $\bullet$ & $\bullet$ & $\bullet$ &  81.08    &   \textbf{20.32}    &  \textbf{50.14}    &   \textbf{55.11 }   \\  \bottomrule
\end{tabular}
\label{tab:moda}
\end{table*}

\begin{figure}[t]
\centering 
 \subfloat{\includegraphics[width=0.40\linewidth]{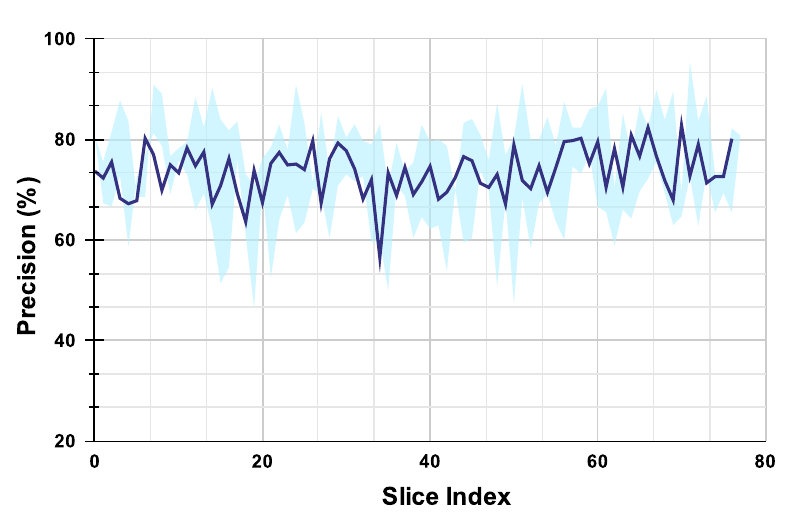}}
 \hspace{2em}
 \subfloat{\includegraphics[width=0.39\linewidth]{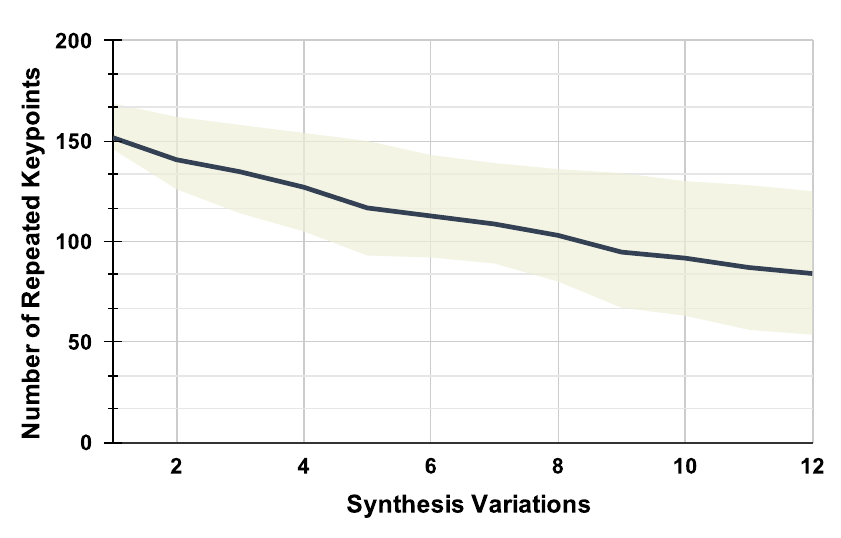}}
 \caption{Repeatability of matches over slices (left) and textures changes (right).}
\label{fig:rep-inv}
\end{figure}

\subsubsection{Ablation study and Evaluation.}
We first measure the impact of varying modalities during US synthesis on the matching performance by excluding and including T1 and Flair modalities. 
It can be observed in Table~\ref{tab:moda} that while the matching precision is only marginally impacted, using all modalities in the synthesis improves the matching scores by more than 12\%. 
In addition, it increases the number of matched points and the percentage of covered area. 
This can be explained by the fact that modalities complete each other when information is missing which highlights the benefits of using our synthesis strategy. 
We also measure the performance of the descriptor across slices, expressed as the average amount by which each matching precision per slice differs from the mean overall volume.
We can observe from Figure~\ref{fig:rep-inv}-left a quasi-constant trend line of matching precision over slices, with a low average standard deviation of 7.1\%.  
Moreover, to measure the repeatability of the descriptor across multiple synthesis modes, we exclude 12 modes from the training set on which we test our model
and calculate the number of repeated keypoints. 
The Figure~\ref{fig:rep-inv}-right shows that more than 50\% of the initially matched points are repeatedly found despite varying the synthesis modes highlighting the texture-invariance properties of the descriptor.   
\begin{figure}[h!]
\centering 
 \includegraphics[width=1\linewidth]{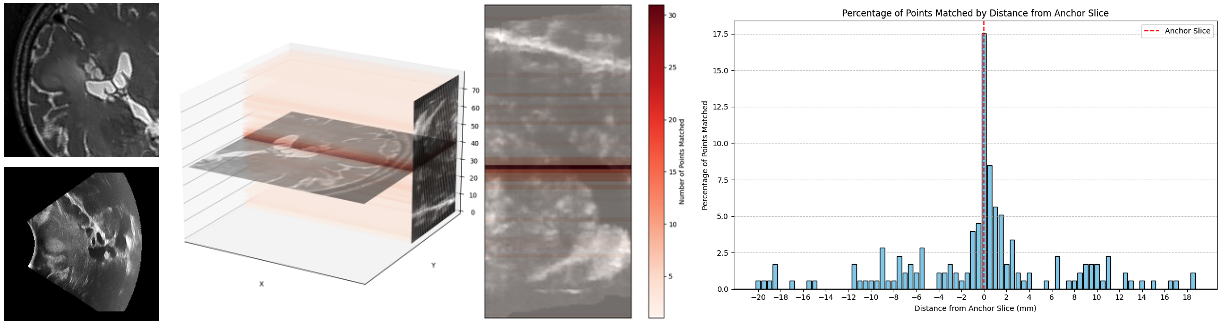}
 \caption{MR slice \#40 retrieved in US volume using descriptor matching.}
\label{fig:search}
\end{figure}

We also perform a slice retrieval test to measure the discriminating properties of our method. 
We search for a target slice over the whole volume by matching $n=200$ keypoints from the target MR slice with $m \times d$ keypoints of the whole US volume.
Results reported in Figure~\ref{fig:search} show that our descriptor can successfully retrieve the target slice and discriminate it over other slices with an average $1.34$\,mm error within 20 slices and  $2.48$\,mm error within 40 slices.

\subsubsection{State-of-the-art comparison.} 
To evaluate the performance of our model against existing image methods, we compared it to three approaches: SIFT \cite{lowe2004distinctive}, which remains the standard for keypoints matching, SuperPoint (SP) \cite{detone2018superpoint} built using a self-supervised learning approach and MIND \cite{heinrich2012mind}, a modality-invariant descriptor for medical imaging, that although not designed for 1-to-1 keypoint matching, is extensively used for multimodal medical image registration through grid regularizing.
We use SIFT and SP as keypoints detectors and descriptors, while we combine MIND with SP keypoints since it only provides a descriptor.
We match these descriptors using both Cosine similarity and the deep neural network LightGlue (LG) \cite{lindenberger2023lightglue} when possible  (SP and Ours).
Results reported in Table~\ref{tab:sota} and shown in Fig.~\ref{fig:matching} show that our approach outperforms these methods in terms of matching score, precision, and number of matched points. 
We only report results on three cases for readability reasons.
Associating our descriptor with Cosine and LG reached similar performance depending on the metric.

\def\arraystretch{0.8}
\rowcolors{3}{}{gray!10}
\begin{table*}[t]
\centering
\caption{Validation on synthetic data and comparisons using real data.}
\begin{tabular}{l | c | c | c | c | c | c | c | c | c }
\toprule
 & \multicolumn{3}{p{9em}|}{\centering Case 1} & \multicolumn{3}{p{9em}|}{\centering Case 2}  & \multicolumn{3}{p{9em}}{\centering Case 3} \\
 \midrule
 Method \hfill &  Prec. \hfill  &  MSc. \hfill  & MP \hfill  & Prec. \hfill & MSc.  \hfill  &  MP \hfill  & Prec. \hfill  & MSc.  \hfill  &  MP \hfill  \\
 \midrule
 SIFT+Cosine  &   4.77    &   3.65    &  153.7    &   2.91    &  2.65     &  182.5  &   2.09    &   2.35    & 224.03 \\
 MIND+Cosine  &   5.15    &   5.15   &   200.0   &   4.44    &  4.44     &  200  &   4.66    &   4.66   & 200 \\  
 SP+Cosine    &   4.58    &  3.45    &  150.8    &   2.76    &   1.58   &  144.83  &   3.82   &   2.80    &  147.58 \\  
 SP+LG        &   55.31    &   10.92  &   25.28   &   13.06    &   1.38    &  13.60  &   17.99    &   2.29    &  16.34 \\  
 Ours+LG   &    80.61   &   \textbf{20.86}    &  51.78    &   \textbf{76.92}    &   4.59    &  11.76  &    53.46   &   3.67   &  13.74 \\ 
 Ours+Cosine &  \textbf{81.08}  &  20.32  &   50.14   &  73.42   &  \textbf{16.38}   &  44.64  &   \textbf{66.35}  &  \textbf{17.90}  & 53.96 \\  
 \bottomrule
\end{tabular}
\label{tab:sota}
\end{table*}

%% file: main.bbl
\begin{thebibliography}{10}
\providecommand{\url}[1]{\texttt{#1}}
\providecommand{\urlprefix}{URL }
\providecommand{\doi}[1]{https://doi.org/#1}

\bibitem{baruch2021joint}
Baruch, E.B., Keller, Y.: Joint detection and matching of feature points in multimodal images. IEEE Transactions on Pattern Analysis and Machine Intelligence  \textbf{44}(10),  6585--6593 (2021)

\bibitem{christy2014retinal}
Christy, D., Moses, C.J.: Retinal image registration feature descriptors-a survey. In: 2014 International Conference on Electronics and Communication Systems (ICECS). pp.~1--5. IEEE (2014)

\bibitem{detone2018superpoint}
DeTone, D., Malisiewicz, T., Rabinovich, A.: Superpoint: Self-supervised interest point detection and description. In: Proceedings of the IEEE conference on computer vision and pattern recognition workshops. pp. 224--236 (2018)

\bibitem{dixon2022intraoperative}
Dixon, L., Lim, A., Grech-Sollars, M., Nandi, D., Camp, S.: Intraoperative ultrasound in brain tumor surgery: a review and implementation guide. Neurosurgical Review  \textbf{45}(4),  2503--2515 (2022)

\bibitem{dorent2023unified}
Dorent, R., Haouchine, N., Kogl, F., Joutard, S., Juvekar, P., Torio, E., Golby, A.J., Ourselin, S., Frisken, S., Vercauteren, T., et~al.: Unified brain mr-ultrasound synthesis using multi-modal hierarchical representations. In: International conference on medical image computing and computer-assisted intervention. pp. 448--458. Springer (2023)

\bibitem{evan2021keymorph}
Evan, M.Y., Wang, A.Q., Dalca, A.V., Sabuncu, M.R.: Keymorph: Robust multi-modal affine registration via unsupervised keypoint detection. In: Medical Imaging with Deep Learning (2021)

\bibitem{ferrante2017slice}
Ferrante, E., Paragios, N.: Slice-to-volume medical image registration: A survey. Medical image analysis  \textbf{39},  101--123 (2017)

\bibitem{Gonzalez-Darder2019}
Gonzalez-Darder, J.M.: State of the Art of the Craniotomy in the Early Twenty-First Century and Future Development, pp. 421--427. Springer International Publishing, Cham (2019)

\bibitem{Haouchine2022}
Haouchine, N., Juvekar, P., Nercessian, M., Wells~III, W.M., Golby, A., Frisken, S.: Pose estimation and non-rigid registration for augmented reality during neurosurgery. IEEE Transactions on Biomedical Engineering  \textbf{69}(4),  1310--1317 (2022)

\bibitem{heinrich2012mind}
Heinrich, M.P., Jenkinson, M., Bhushan, M., Matin, T., Gleeson, F.V., Brady, M., Schnabel, J.A.: Mind: Modality independent neighbourhood descriptor for multi-modal deformable registration. Medical image analysis  \textbf{16}(7),  1423--1435 (2012)

\bibitem{jiang2021review}
Jiang, X., Ma, J., Xiao, G., Shao, Z., Guo, X.: A review of multimodal image matching: Methods and applications. Information Fusion  \textbf{73},  22--71 (2021)

\bibitem{joutard2022driving}
Joutard, S., Dorent, R., Ourselin, S., Vercauteren, T., Modat, M.: Driving points prediction for abdominal probabilistic registration. In: International Workshop on Machine Learning in Medical Imaging. pp. 288--297. Springer (2022)

\bibitem{juvekar2023remind}
Juvekar, P., Dorent, R., K{\"o}gl, F., Torio, E., Barr, C., Rigolo, L., Galvin, C., Jowkar, N., Kazi, A., Haouchine, N., et~al.: Remind: The brain resection multimodal imaging database. medRxiv  (2023)

\bibitem{kumar2013content}
Kumar, A., Kim, J., Cai, W., Fulham, M., Feng, D.: Content-based medical image retrieval: a survey of applications to multidimensional and multimodality data. Journal of digital imaging  \textbf{26},  1025--1039 (2013)

\bibitem{lindenberger2023lightglue}
Lindenberger, P., Sarlin, P.E., Pollefeys, M.: Lightglue: Local feature matching at light speed. arXiv preprint arXiv:2306.13643  (2023)

\bibitem{lowe2004distinctive}
Lowe, D.G.: Distinctive image features from scale-invariant keypoints. International journal of computer vision  \textbf{60},  91--110 (2004)

\bibitem{JIE}
Luo, J., Toews, M., Machado, I., Frisken, S., Zhang, M., Preiswerk, F., Sedghi, A., Ding, H., Pieper, S., Golland, P., Golby, A., Sugiyama, M., Wells~III, W.M.: A feature-driven active framework for ultrasound-based brain shift compensation. In: MICCAI 2018. pp. 30--38 (2018)

\bibitem{Machado}
Machado, I., Toews, M., Luo, J., Unadkat, P., Essayed, W., George, E., Teodoro, P., Carvalho, H., Martins, J., Golland, P., Pieper, S., Frisken, S., Golby, A., III, W.: Non-rigid registration of 3d ultrasound for neurosurgery using automatic feature detection and matching. International Journal of Computer Assisted Radiology and Surgery  \textbf{13} (06 2018)

\bibitem{paulus2017handling}
Paulus, C.J., Haouchine, N., Kong, S.H., Soares, R.V., Cazier, D., Cotin, S.: Handling topological changes during elastic registration: Application to augmented reality in laparoscopic surgery. International journal of computer assisted radiology and surgery  \textbf{12},  461--470 (2017)

\bibitem{Sanai}
Sanai, N., Polley, M.Y., McDermott, M.W., Parsa, A.T., Berger, M.S.: An extent of resection threshold for newly diagnosed glioblastomas: Clinical article. Journal of Neurosurgery JNS  \textbf{115}(1), ~3--8 (2011)

\bibitem{talbot2015}
Talbot, H., Haouchine, N., Peterlik, I., Dequidt, J., Duriez, C., Delingette, H., Cotin, S.: {Surgery Training, Planning and Guidance Using the SOFA Framework}. In: Hege, H.C., Ropinski, T. (eds.) Eurographics 2015 - Dirk Bartz Prize. The Eurographics Association (2015). \doi{10.2312/egm.20151028}

\bibitem{verdie2015tilde}
Verdie, Y., Yi, K., Fua, P., Lepetit, V.: Tilde: A temporally invariant learned detector. In: Proceedings of the IEEE conference on computer vision and pattern recognition. pp. 5279--5288 (2015)

\bibitem{wu2018multimodal}
Wu, M., Goodman, N.: {Multimodal Generative Models for Scalable Weakly-Supervised Learning}. {{NeurIPS}}  \textbf{31} (2018)

\end{thebibliography}
